\documentclass[journal,onecolumn]{IEEEtran}

\ifCLASSINFOpdf
\else
   \usepackage[dvips]{graphicx}
\fi
\usepackage{url}

\usepackage{graphicx}
\usepackage{amsmath}
\usepackage{amssymb}
\usepackage{xcolor}
\usepackage{cite}
\usepackage{xcolor}
\usepackage[linesnumbered,ruled,vlined]{algorithm2e}

\SetCommentSty{mycommfont}
\SetKwInput{KwInput}{Input}                % Set the Input
\SetKwInput{KwOutput}{Output}              % set the Output

\begin{document}
	
	\begin{center}
		{\Large \textbf{Using MM principles to deal with incomplete data in K-means clustering}}\\
		\vspace{1em}
		{\large Mini Project: MM Optimization Algorithms}\\
		\vspace{1em}
		{\large Ali Beikmohammadi}\\
		\vspace{1em}
		\textit{Department of Computer and Systems Sciences\\
  Stockholm University\\
  SE-164 07 Kista, Sweden \\
  \texttt{beikmohammadi@dsv.su.se} \\}
	\end{center}

	\begin{center}
		\rule{\textwidth}{0.2mm}
		\vspace{1mm}
	\end{center}		

	\begin{abstract}
Among many clustering algorithms, the K-means clustering algorithm is widely used because of its simple algorithm and fast convergence. However, this algorithm suffers from incomplete data, where some samples have missed some of their attributes. To solve this problem, we mainly apply MM principles to restore the symmetry of the data, so that K-means could work well. We give the pseudo-code of the algorithm and use the standard datasets for experimental verification. The source code for the experiments is publicly available in the following link: \url{https://github.com/AliBeikmohammadi/MM-Optimization/blob/main/mini-project/MM%20K-means.ipynb}.

	%\textbf{Collaborators}: list the names and affiliations of expected collaborators on the project here
	\end{abstract}

	\begin{center}
		\rule{\textwidth}{0.2mm}
	\end{center}		

	\vspace{5mm}
	
%\begin{multicols*}{2}

\section{Background and Introduction}
Clustering is the task of grouping a set of objects in such a way that objects in the same group (called a cluster) are more similar (in some sense or another) to each other than to those in other groups (clusters). It is the main task of exploratory data mining, and a common technique for statistical data analysis used in many fields, including machine learning, pattern recognition, image analysis, information retrieval, and bioinformatics \cite{boyd2018introduction, jain1999data, bijuraj2013clustering}.

From a theoretical point of view, all proposed methods of clustering could be appointed to five major classes: Partitioning methods, Hierarchical methods, Density-based methods, Grid-based methods, and Model-based methods. Particularly, the simplest form of clustering is partitional clustering which aims at subdividing the dataset into a set of K groups so that specific clustering criteria are optimized, where K is the number of groups pre-specified by the analyst \cite{likas2003global}. The most widely used criterion is the clustering error criterion which for each point computes its squared distance from the corresponding cluster center and then takes the sum of these distances for all points in the data set \cite{likas2003global, bishop2006pattern, webb2011statistical}. 

There are different types of partitioning clustering methods. A popular clustering method that minimizes the clustering error is the K-means algorithm \cite{macqueen1967some}, in which, each cluster is represented by the center or means of the data points belonging to the cluster. 

K-means algorithm has many advantages such as simple mathematical ideas, fast convergence, easy implementation, and easily adaptation to new examples. \cite{li2017parallel}. Therefore, the application fields are very broad, including different types of document classification, topic discovery, patient clustering, customer market segmentation, student clustering, and weather zones, the construction of recommendation systems based on user interests, and so on \cite{boyd2018introduction, li2017parallel}. 
However, the K-means algorithm is a local search procedure and it is well known that it suffers from the serious drawback such as, choosing K manually, being dependent on initial values, difficulty in clustering data of varying sizes and density, sensitivity to outliers, scaling with the number of dimensions, and the need for access to full data \cite{pena1999empirical}. Fortunately, the researchers have proposed some solutions to address many of these challenges \cite{yuan2019research, kanungoy2000cient, huang1998extensions}. 

In this report, we will focus on the last weakness, where we have an incomplete data. The assumption of not having access to all the attributes of a sample is completely realistic. For example, in many cases where these components of each sample are collected from the environment through various sensors, there is always the possibility that some of the sensors will fail. Under this condition, we would not have access to those features for that particular sample. The first idea when dealing with such incomplete samples is to withdraw them from the data set, so that it is possible to use a standard K-means clustering algorithm, known as Lloyd's algorithm \cite{lloyd1982least}. But, in this project, with the help of MM optimization principles \cite{Kenneth-2016}, the symmetry of the dataset is reconstructed. By doing this, on the one hand, we make the most of the available measurements, and on the other hand, we are able to use the standard K-means clustering algorithm.

The rest of this work is organized as follows. In Section \ref{section2}, we  describe the formulation associated with the original problem which is K-means clustering algorithm. We  apply MM principles to design our solution in Section \ref{section3} to deal with incomplete data. Experimental results and analyses are reported in Sections \ref{section4}.

\section{Problem Formulation: K-means Clustering Algorithm} \label{section2}
The K-means clustering algorithm was first proposed in 1957 by Stuart Lloyd, and independently by Hugo Steinhaus \cite{lloyd1982least, boyd2018introduction}. Sometimes, it is  called the Lloyd algorithm. The name `K-means' has been used from the 1960s. 

The K-means algorithm is based on alternating two procedures. The first is one of assignment of objects to groups. An object is usually assigned to the group to whose mean it is closest in the Euclidean sense. The second procedure is the calculation of new group means based on the assignments. The process terminates when no movement of an object to another group will reduce the within-group sum of squares.

To be more specific, given a set of observations $(x_1, x_2, ..., x_m)$, where each observation is a $d$-dimensional real vector, K-means clustering aims to partition the $m$ observations into $K (\le m)$ sets $C = \{C_1, C_2, ..., C_K\}$ so as to minimize the within-cluster sum of squares. Formally, the objective is to:

\begin{equation}
\label{eq:1}
    \text{minimize} f(\mu) = \sum_{k=1}^{K} \sum_{x_i \in C_k} \|x_i-\mu_k\|^2,
\end{equation}
where $\mu_k$ is the center of cluster $k$. The solution to the centroid $\mu_k$ is as follows:
\begin{equation}
\label{eq:2}
\begin{split}
\frac {\partial f(\mu)} { \partial\mu_k} &= \frac {\partial} {\partial\mu_k} \sum_{k=1}^{K} \sum_{x_i \in C_k} (x_i-\mu_k)^2\\
&=  \sum_{k=1}^{K} \sum_{x_i \in C_k}\frac {\partial} {\partial\mu_k} (x_i-\mu_k)^2\\
&=  \sum_{x_i \in C_k} 2 (x_i-\mu_k).
\end{split}
\end{equation}

Let Equation \ref{eq:2} be zero; then $\mu^\star_k=\frac {1} { |C_k|}\sum_{x_i \in C_k}x_i$. 

In terms of iterative solving this problem, the Algorithm \ref{code1} shows all the needed steps, where the central idea is to randomly extract K sample points from the sample set as the center of the initial cluster; Divide each sample point into the cluster represented by the nearest center point; then the center point of all sample points in each cluster is the center point of the cluster. Repeat the above steps until the center point of the cluster is almost unchanged ($\le \epsilon$) or reaches the set number of iterations.

\begin{minipage}{0.5\textwidth}
\vspace{-63pt}
\centering
\begin{algorithm}[H] \label{code1}
\DontPrintSemicolon
  
  \KwInput{$K, \epsilon$}
  \KwOutput{$\mu_1, \mu_2, ... , \mu_K$}
  \KwData{$X$}
  $n=0$ \;
  %Randomly initialize K centroids: $\mu_1^n, \mu_2^n, ... , \mu_K^n \in \mathbb{R}^d$ \;
  Randomly pick K samples of X as K centroids: $\mu_1^n, \mu_2^n, ... , \mu_K^n \in \mathbb{R}^d$ \;
 \Repeat{$\sum_{k=1}^{K}\|\mu_k^n-\mu_k^{n-1}\|^2 \le \epsilon$}{
    $n \leftarrow n+1$\;
    $C_k \leftarrow \emptyset $ for all $k=1, ... , K$\;
    \tcc{Cluster Assignment Step}
    \ForEach{$x_i \in X$}{
    $k^\star\leftarrow \text{arg min}_k \|x_i-\mu_k^n\|^2$ \tcc{Assign $x_i$ to closest centroid}
    $C_{k^\star}\leftarrow C_{k^\star} \cup x_i$
    }
    \tcc{Centroid Update Step} 
    \ForEach{$k=1 $ to $ K$}{
    $\mu_k^n=\frac {1} { |C_k|}\sum_{x_i \in C_k}x_i$}
    }
\caption{K-means}
\end{algorithm}
\end{minipage}
\hfill
\begin{minipage}{0.5\textwidth}
\centering
\begin{algorithm}[H] \label{code2}
\DontPrintSemicolon
  
  \KwInput{$K, \epsilon$}
  \KwOutput{$\mu_1, \mu_2, ... , \mu_K$}
  \KwData{$X, O_X$}
  $n=0$ \;
  %Randomly initialize K centroids: $\mu_1^n, \mu_2^n, ... , \mu_K^n \in \mathbb{R}^d$ \;
  Randomly pick K samples among those X that are fully observed as K centroids: $\mu_1^n, \mu_2^n, ... , \mu_K^n \in \mathbb{R}^d$ \;
  Randomly replace unobserved data with corresponding element of one of $\mu_1^n, \mu_2^n, ... , \mu_K^n$  \;
 \Repeat{$\sum_{k=1}^{K}\|\mu_k^n-\mu_k^{n-1}\|^2 \le \epsilon$}{
    $n \leftarrow n+1$\;
    $C_k \leftarrow \emptyset $ for all $k=1, ... , K$\;
    \tcc{Cluster Assignment Step}
    \ForEach{$x_i \in X$}{
    $k^\star\leftarrow \text{arg min}_k \|x_i-\mu_k^n\|^2$ \tcc{Assign $x_i$ to closest centroid}
    $C_{k^\star}\leftarrow C_{k^\star} \cup x_i$
    }
    \tcc{Centroid Update Step} 
    \ForEach{$k=1 $ to $ K$}{
    $\mu_k^n=\frac {1} { |C_k|}\sum_{x_i \in C_k}x_i$}
    \tcc{Unobserved Data Replacement Step} 
    \ForEach{$x_i \in X$ that $j \notin O_{x_i}$}{
    ${x_i}_j \leftarrow {\mu^n_k}_j$\;
    }
    }
\caption{MM K-means}
\end{algorithm}
\end{minipage}

However, as it turns out, this algorithm can only be used when all d features of each observation are available. In the next section, we overcome this limitation by applying MM principles.

\section{Restoring Data Symmetry by Applying MM Principles} \label{section3}
Let $O_{x_i}$ denote the set of indexes observed in sample $x_i$. Then, the objective function mentioned in Equation \ref{eq:1} should change to
\begin{equation}
\label{eq:3}
    \text{minimize} f(\mu) = \sum_{k=1}^{K} \sum_{x_i \in C_k} \Big[\sum_{j \in O_{x_i}}({x_i}_j-{\mu_k}_j)^2\Big],
\end{equation}
due to incomplete $x_i$s. Note that, ${\mu_k}_j$ is $j$th index of $\mu_k$. Therefore, incomplete data prevents the use of the standard K-means algorithm and following Algorithm \ref{code1}.

However, since $f(\mu)$ is a convex function, we can apply an interesting majorization to $f(\mu)$ by following the MM principles so that:
\begin{equation}
\label{eq:4}
    f(\mu) \le \sum_{k=1}^{K} \sum_{x_i \in C_k} \Big[\sum_{j \in O_{x_i}}({x_i}_j-{\mu_k}_j)^2 + \sum_{j \notin O_{x_i}}({\mu^n_k}_j-{\mu_k}_j)^2\Big]
    =g(\mu|\mu^n).
\end{equation}

As we know, majorization combines two conditions: the tangency condition $g(\mu^n|\mu^n)= f(\mu^n)$ and the domination condition $g(\mu|\mu^n)\ge f(\mu)$ for all $\mu$ \cite{Kenneth-2016}. Here, since $\sum_{j \notin O_{x_i}}({\mu^n_k}_j-{\mu^n_k}_j)^2 =0$ and $\sum_{j \notin O_{x_i}}({\mu^n_k}_j-{\mu_k}_j)^2 \ge 0$, both conditions are satisfied. 

Therefore, by simply substituting the corresponding ${\mu^n_k}_j$ with the unobserved data ${x_i}_j$ in each iteration, the data symmetry is restored, and one can apply standard K-means algorithm. Particularly, Algorithm \ref{code2} shows a step-by-step process of our proposed method. In the next section, we examine its performance in detail.
	
\section{Results and Discussion} \label{section4}
\begin{figure}[t]
    \centering
	\includegraphics[width=1\columnwidth]{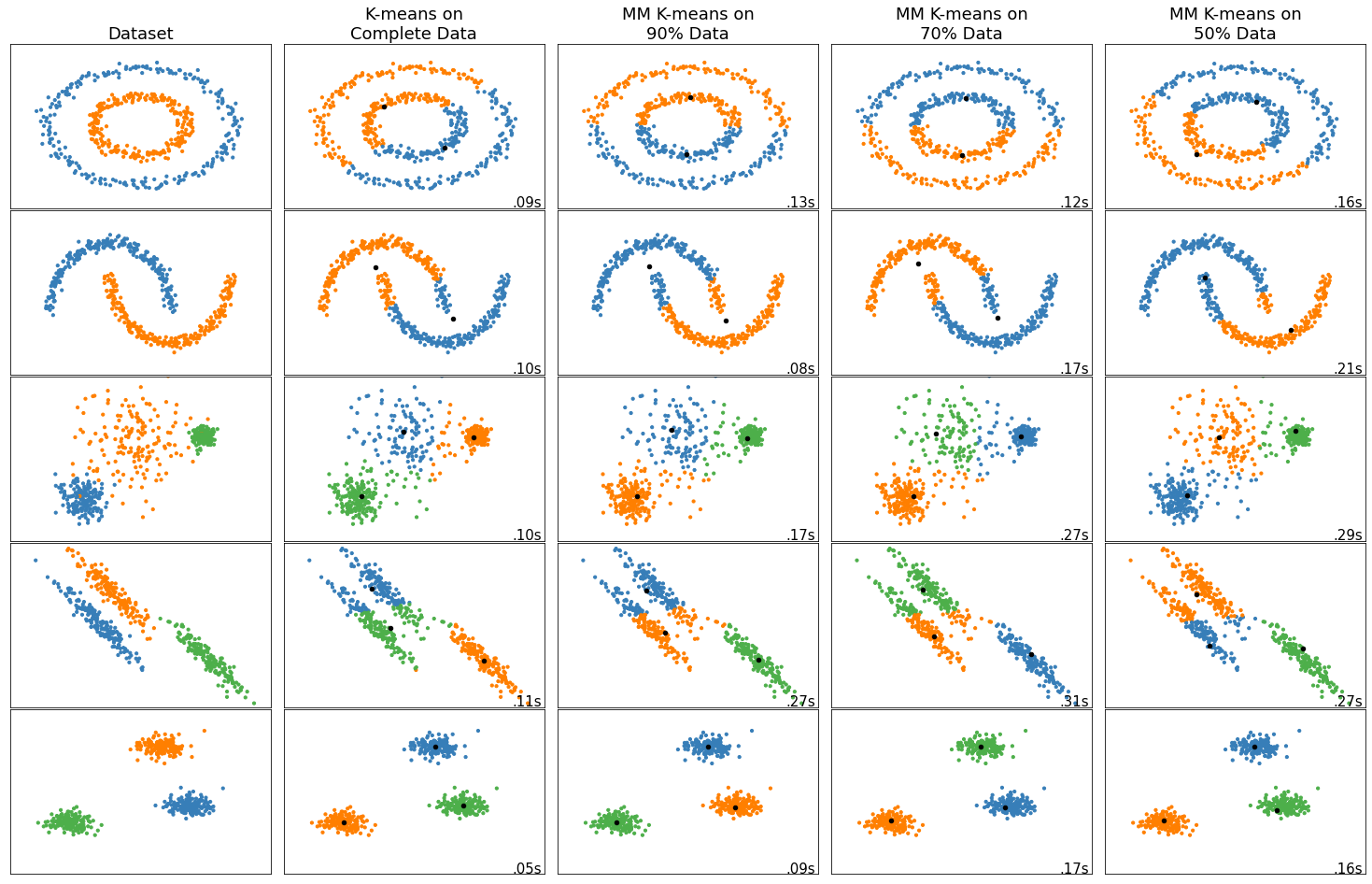}
	\caption{Comparison of the K-means algorithm with the proposed MM K-means algorithm when 10\%, 30\% and 50\% of the data are unobserved randomly. The first column represents noisy circles, noisy moons, blobs with varied variances, anisotropicly distributed, and blobs dataset, respectively. Note that the black dots in each plot represent the cluster centers reached by each algorithm after 100 iterations.}
	\label{fig:fig1}
\end{figure}

\begin{figure}[t]
    \centering
	\includegraphics[width=1\columnwidth]{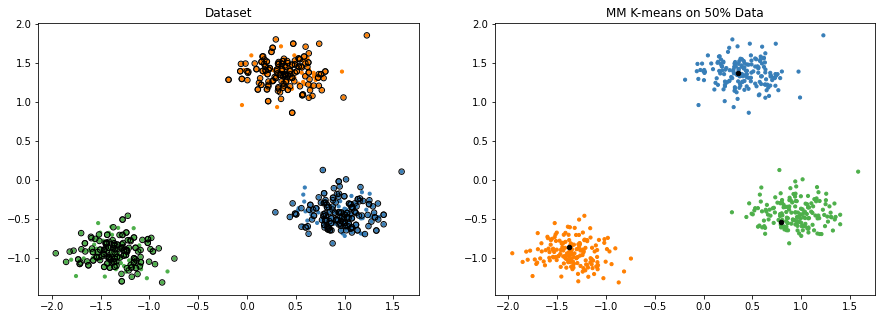}
	\caption{(Left) Blobs dataset with three clusters of green, blue, and orange, where samples that have lost at least one of their elements are marked with the black circle. Note that, here, 50\% of all elements, i.e., 50\% of 500*2, are assumed unobserved. (Right) The result of applying the proposed MM K-means Algorithm to the Blobs dataset (left plot) in more detail, where the centers of the clusters are marked with black dots.}
	\label{fig:fig2}
\end{figure}

For comparison, we have created five different datasets shown in the first column of Figure \ref{fig:fig1} using the scikit-learn library \cite{pedregosa2011scikit}. To make good visualization, each sample is assumed to consist of only two elements, i.e., $x_i \in \mathbb{R}^2$ \$. We have also collected 500 samples in each dataset. Also, the first two datasets, namely noisy circles and noisy moons, consist of two clusters. Note that, when making them, the amount of noise is set to 0.05. The rest of the datasets, blobs with varied variances, anisotropicly distributed, and blobs dataset,comprised of three clusters. Details on how to create datasets can be found in the source code in the following link: \url{https://github.com/AliBeikmohammadi/MM-Optimization/blob/main/mini-project/MM%20K-means.ipynb}.

\begin{table*}
	\centering
	\begin{tabular}{ccccccccc}
		\hline
		\hline
		\textbf{Dataset} & \textbf{Algorithm} & \textbf{Time} & \textbf{Homogeneity}& \textbf{Completeness}& \textbf{V-measure}& \textbf{ARI}& \textbf{AMI}& \textbf{Silhouette Coefficient} \\
		\hline
		\hline
		    noisy circles & original dataset & 0.00	& 1.000	& 1.000 &	1.000 &	1.000 &	1.000 &	0.111 \\
        		 &	K-means on Complete Data &	0.09 &	0.000 &	0.000&	0.000&	-0.002	&-0.001	&0.352\\
        		 &	MM K-means on 90\% Data &  	0.13&	0.000&	0.000&	0.000&	-0.002&	-0.001&	0.355\\
                &	MM K-means on 70\% Data  &	0.12&	0.000&	0.000&	0.000&	-0.002&	-0.001&	0.355\\
                &	MM K-means on 50\% Data  &	0.16&	0.000&	0.000&	0.000&	-0.002&	-0.001&	0.347\\
              
                \hline
noisy moons&	original dataset                 &0.00&	1.000&	1.000&	1.000&	1.000&	1.000&	0.387\\
                &	K-means on Complete Data&	0.10&	0.385&	0.385&	0.385&	0.483&	0.384&	0.497\\
                &	MM K-means on 90\% Data  &	0.08&	0.385&	0.385&	0.385&	0.483&	0.384&	0.497\\
                &	MM K-means on 70\% Data  &	0.17&	0.386&	0.386&	0.386&	0.483&	0.385&	0.495\\
                &	MM K-means on 50\% Data  &	0.21&	0.387&	0.394&	0.391&	0.467&	0.390&	0.484\\
                \hline
varied &  	original dataset                & 	0.00&	1.000&	1.000&	1.000&	1.000&	1.000&	0.569\\
              & 	K-means on Complete Data&	0.10&	0.726&	0.743&	0.734&	0.731&	0.733&	0.639\\
              & 	MM K-means on 90\% Data  &	0.17&	0.723&	0.740&	0.731&	0.727&	0.730&	0.639\\
              & 	MM K-means on 70\% Data  &	0.27&	0.702&	0.723&	0.712&	0.701&	0.711&	0.631\\
              & 	MM K-means on 50\% Data  &	0.29&	0.737&	0.752&	0.745&	0.745&	0.744&	0.635\\
              \hline
aniso  &  	original dataset                 &	0.00&	1.000&	1.000&	1.000&	1.000&	1.000&	0.459\\
               & 	K-means on Complete Data&	0.11&	0.600&	0.600&	0.600&	0.578&	0.599&	0.503\\
               & 	MM K-means on 90\% Data  &	0.27&	0.613&	0.615&	0.614&	0.585&	0.613&	0.503\\
               & 	MM K-means on 70\% Data  &	0.31&	0.642&	0.647&	0.645&	0.618&	0.643&	0.504\\
               & 	MM K-means on 50\% Data  &	0.27&	0.657&	0.680&	0.668&	0.619&	0.667&	0.490\\
               \hline
blobs   & 	original dataset                 &	0.00&	1.000&	1.000&	1.000&	1.000&	1.000&	0.829\\
               & 	K-means on Complete Data&	0.05&	1.000&	1.000&	1.000&	1.000&	1.000&	0.829\\
               & 	MM K-means on 90\% Data  &	0.09&	1.000&	1.000&	1.000&	1.000&	1.000&	0.829\\
               & 	MM K-means on 70\% Data  &	0.17&	1.000&	1.000&	1.000&	1.000&	1.000&	0.829\\
               & 	MM K-means on 50\% Data  &	0.16&	1.000&	1.000&	1.000&	1.000&	1.000&	0.829\\
		\hline
		\hline
	\end{tabular}
	\caption{Numerical comparison of the K-means algorithm with the proposed MM K-means algorithm when 10\%, 30\% and 50\% of the data are unobserved randomly. Note that, ARI and AMI stand for Adjusted Rand Index and Adjusted Mutual Information, respectively. Also, anisotropicly distributed dataset and blobs with varied variances dataset are mentioned with aniso and varied, respectively.}
	\label{tbl:table1}
\end{table*}

Four experiments were performed on each dataset. First, the K-means algorithm is trained on the whole dataset as the baseline. Second, after removing 10\% of the total constituent elements of the samples, the proposed MM K-means algorithm is tested. The third and fourth are done exactly as same as the second experiment, with the difference that 30\% and 50\% of the data are assumed not to be observed, respectively. In all experiments, the number of iterations is set to 100.

Following the algorithms \ref{code1} and \ref{code2}, the results shown in Figure \ref{fig:fig1} are obtained. These results visually confirm that the proposed MM K-means algorithm is fully consistent with the K-means algorithm and was able to restore the data symmetry. Specifically, the black dots shown in each plot in Figure \ref{fig:fig1}, which represent the centers of each cluster, indicate that both algorithms have converged to the same point. But, as expected, these found centers can not always guarantee the finding of perfectly correct clusters. In fact, we were able to use MM principle to help reconstruct the data, not to improve the K-means algorithm. However, it can be argued that the same principle can be used as an umbrella over other more efficient clustering methods that suffer from the inability to work with missing data.

Interestingly, the proposed method is robust against increasing the amount of missing data, where by increasing the amount of unobserved data from 10\% to 50\%, it has a negligible effect on the results. To prove this claim more, Table \ref{tbl:table1} provides a numerical comparison with full details. In this comparison, known criteria in the field of clustering have been used, which are: Homogeneity, Completeness, V-measure, Adjusted Rand Index, Adjusted Mutual Information, and Silhouette Coefficient. Regardless of the criteria chosen, all the results confirm that, firstly, the data reconstruction is very well done and, secondly, the increase in the amount of lost data has little effect on the performance.
Figure \ref{fig:fig2} also shows the proper scattering of incomplete data (assuming 50\% of incomplete data) in the dataset. However, it is clear that the algorithm has been able to find cluster centers similar to the ones in which we have access to all data, by properly reconstructing missing data. Finally, as shown in Figure \ref{fig:fig1} and Table \ref{tbl:table1}, the execution time of the proposed MM K-means algorithm is longer because it requires updating the missing elements in each iteration. However, in conclusion, applying MM principles to reconstruct data symmetry can play an important role in maximizing data usage along with the possibility of using standard algorithms. It is hoped that in the future, this technique can be used to improve the performance of deep learning-based methods for various applications such as plant identification \cite{beikmohammadi2020swp, BEIKMOHAMMADI2022117470}, handwritten digit recognition \cite{beikmohammadi2021hierarchical}, and human action detection \cite{beikmohammadi2019mixture}.

\bibliographystyle{IEEEtran}
\bibliography{./Mini-Project-MM-Optimization.bib}
\end{document}